\documentclass[sigconf,screen,review=false,anonymous=false]{acmart}
\acmConference[SC'25]{The International Conference for High Performance
Computing, Networking, Storage, and Analysis}{November 16–21}{St. Louis, MO}
\acmYear{2025}
\acmISBN{} 
\acmDOI{} 
\startPage{1}

\setcopyright{none}

\AtBeginDocument{%
  }

\usepackage{algorithm,algorithmic,amsfonts,epigraph,fancyvrb,fixme,fontenc,graphicx,soul,subcaption,xcolor}
\usepackage{url}
\usepackage{hyperref}
\usepackage[final]{listings}
\usepackage[inline]{enumitem}
\fxsetup{status=final, theme=color}
\fxsetface{margin}{\tiny}
\VerbatimFootnotes
\setlist{nosep}
\graphicspath{{figs}}
\definecolor{codegreen}{rgb}{0,0.6,0}
\definecolor{codegray}{rgb}{0.5,0.5,0.5}
\definecolor{codepurple}{rgb}{0.58,0,0.82}
\newcommand{\TADASHI}[0]{\mbox{\textsc{Tadashi}}}

\begin{document}

\title{\TADASHI{}: Enabling AI-Based Automated Code Generation With Guaranteed Correctness}


\author{Emil Vatai}
\email{emil.vatai@riken.jp}
\orcid{0000-0001-7494-5048}
\affiliation{%
  \institution{RIKEN Center for Computational Science}
  \city{Kobe}
  \state{Hyogo}
  \country{Japan}
}

\author{Aleksandr Drozd}
\email{aleksandr.drozd@riken.jp}
\affiliation{%
  \institution{RIKEN Center for Computational Science}
  \city{Kobe}
  \state{Hyogo}
  \country{Japan}
}

\author{Ivan R. Ivanov}
\email{ivanov.i.e641@m.isct.ac.jp}
\affiliation{%
  \institution{Institute of Science Tokyo / RIKEN Center for Computational Science}
  \city{Kobe}
  \state{Hyogo}
  \country{Japan}
}

\author{Jo\~ao E. Batista}
\email{joao.batista@riken.jp}
\affiliation{%
  \institution{RIKEN Center for Computational Science}
  \city{Kobe}
  \state{Hyogo}
  \country{Japan}
}

\author{Yinghao Ren}
\email{yinghaoren.cs@gmail.com}
\affiliation{%
  \institution{ByteDance inc.}
  \city{Shanghai}
  \country{China}
}

\author{Mohamed Wahib}
\email{mohamed.attia@riken.jp}
\affiliation{%
  \institution{RIKEN Center for Computational Science}
  \city{Kobe}
  \state{Hyogo}
  \country{Japan}
}

\renewcommand{\shortauthors}{Vatai et al.}

\begin{abstract}
Frameworks and domain-specific languages for auto-generating code have traditionally depended on human experts to implement rigorous methods ensuring the legality of code transformations. Recently, machine learning (ML) has gained traction for generating code optimized for specific hardware targets. However, ML approaches-particularly black-box neural networks-offer no guarantees on the correctness or legality of the transformations they produce.
To address this gap, we introduce \TADASHI{}, an end-to-end system that leverages the polyhedral model to support researchers in curating datasets critical for ML-based code generation. \TADASHI{} provides an end-to-end system capable of applying, verifying, and evaluating candidate transformations on polyhedral schedules with both reliability and practicality. We formally prove that \TADASHI{} guarantees the legality of generated transformations, demonstrate its low runtime overhead, and showcase its broad applicability. \TADASHI{} available at \url{https://github.com/vatai/tadashi/}.
\end{abstract}

%

\ccsdesc[500]{Software and its engineering~Source code generation}
\ccsdesc[500]{Software and its engineering~Correctness}
\ccsdesc[500]{Software and its engineering~Software performance}

\ccsdesc[500]{Computer systems organization~Embedded systems}
\ccsdesc[300]{Computer systems organization~Redundancy}
\ccsdesc{Computer systems organization~Robotics}
\ccsdesc[100]{Networks~Network reliability}


\begin{teaserfigure}
  \centering
  \includegraphics{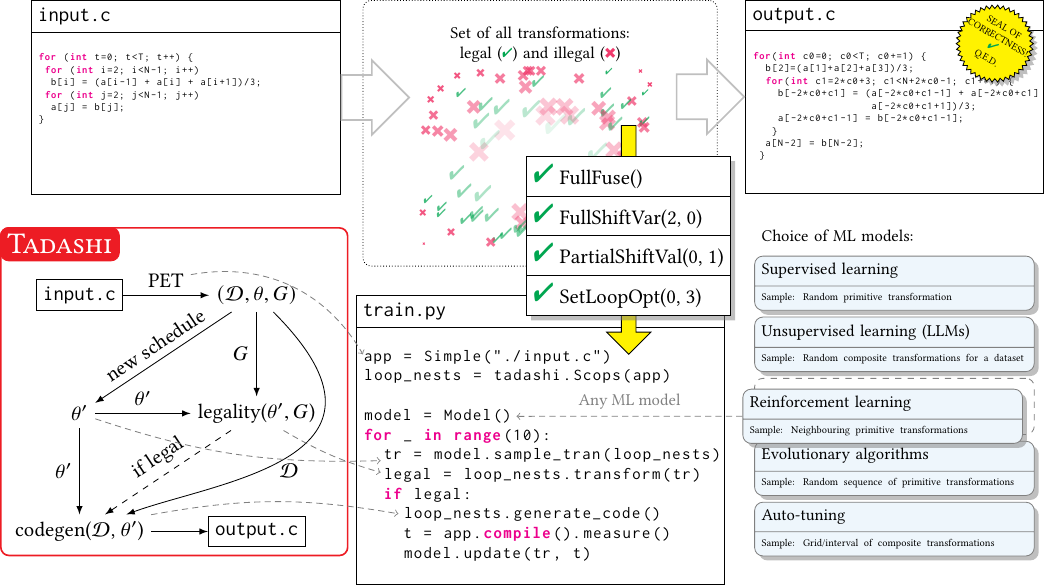}
  \caption{\TADASHI{} provides a simple interface to allow different ML methods to sample from the massive combinatorial space of possible loop transformations, with assurance that the sampled transformation is \emph{legal}. Using the polyhedral model, \TADASHI{} is able to verify whether a transformation is legal or not, and generate the code with the legal transformations.}
  \Description{Tadashi sampling the space of transformations}
  \label{fig:teaser}
\end{teaserfigure}


\maketitle

\section{Introduction}
Machine learning (ML) is disrupting all forms of digital media, and the creation of computer programs is no exception.
ML can be and has been applied in various ways with the aim of making the work of programmers~\cite{dehaerne2022code} easier: generating code from ``natural languages''~\cite{phan2019self}, write a summary (e.g., write commit messages) of a piece of code~\cite{shido2019automatic}, auto-complete code on-the-fly while it's being written~\cite{github2022github} among others. In this paper, we consider the task of optimising code to a hardware target, i.e., using ML to transform the hotspots in the code, i.e., loops, in a way that improves performance.

Checking if the program generated by the ML model is faster then the baseline is just a matter of running it, then comparing against the run-time of the original code. However, it is a major challenge to get formal guarantees that the transformed code produces the same output as the original~\cite{dehaerne2022code,article-Munoz,jiang2024survey}. This impedes the adoption of ML for code generation and transformation in mission critical scientific/engineering codes, or codes that come with high liability, such as in the financial, legal, and corporate sectors. Additionally, there is a lack of tooling for generating and sharing parameterized datasets comprised of legal auto-generated code, which further constraints progress in the field since, to date, ML-based solutions have to curate their datasets.

To this end, we developed \TADASHI{}, a Python library for sampling the search space of possible loop transformations, with the capability to verify the legality (i.e., correctness) of the sampled transformations. \TADASHI{} allows the user to programmatically specify the pattern for sampling legal individual or combinations of transformations, based on the preference of the ML method that would be used to explore the search space of transformations. Through the use of the polyhedral model, \TADASHI{} is capable of enabling ML engineers to explore novel and more aggressive methods of code optimisation in the datasets they curate, while guaranteeing that the results of the generated code are functionally identical to the original program. \TADASHI{} is an end-to-end solution to transform, verify and evaluate transformations, emphasizing the ease-of-use for ML practitioners, with support for HPC oriented libraries such as MPI4Py futures~\cite{rogowski2023mpi4pyfutures}; \TADASHI{} aims to provide a simple Python interface, curating datasets or performing direct code-generation at scale to be used with different ML approaches.

We envision \TADASHI{} with the overarching purpose of helping clusters and supercomputers reach their full potential. Namely, after the deployment of a new (or existing) super\-computing system, one can train an ML model to learn to transform hotspots in scientific software to run optimally on the given hardware. By basing their ML models on \TADASHI{}, end users will have both fully correct and highly optimized code. With this in mind \TADASHI{} would be a viable and practicable option when porting existing science codes to new supercomputers. This is particularly a huge challenge, as demonstrate in recent history by the Exascale Computing Project~\cite{heroux2024scalable,exascale}, which required 7-years, \$1.8 billion, and approximately 1,000 researchers from DOE national laboratories to port the codes to new systems (among other goals of  ECP).


In this paper, we make the following contributions:
\begin{itemize}[leftmargin=2.5mm]
\item We introduce \TADASHI{}, a Python-based end-to-end system designed for ML practitioners seeking to design tools for code auto generation with correctness guarantees. \TADASHI{} lowers the barrier to entry of polyhedral loop transformations; a model that can give mathematically provable guarantees on the correctness (i.e., legality) of the transformed code.
\item We demonstrate how \TADASHI{} can be used by an ML system by implementing two representative ML models on top of \TADASHI{}.
\item We show the capability of \TADASHI{} to speed up software on Polybench suite and two science applications.
\end{itemize}

\section{Background and Motivation}

In this section we discuss the motivation for designing a library that leverages the polyhedral model as bases for guaranteeing correctness of ML-driven code generation. 

\subsection{Correctness in ML-Based Code Auto-Generation}
ML is being increasingly used in code auto-generation, at different levels of abstraction, e.g., source code~\cite{stein2020exploring}, abstract syntax tree (AST)~\cite{shido2019automatic}, polyhedral~\cite{baghdadi2021deepa}, dependency graphs~\cite{cummins2020programl}, intermediate Representations (IR)~\cite{ben-nun2018neural}, and assembly instructions~\cite{mankowitz2023faster}.

To ensure correctness, existing approaches for using ML in code generation or transformation rely on one of the following approaches:
\begin{itemize}[leftmargin=2.5mm]
\item Extensive unit testing: for this, widely used, approach, a suite of unit tests is used to validate correctness~\cite{cummins2024meta,beer2024examinationcodegeneratedlarge,10.1145/3597503.3639219,chen2021codex}. While unit testing can be effective, it provides no guarantees on correctness\footnote{"\textit{Program testing can be used to show the presence of bugs, but never to show their absence!}"-- Edsger W. Dijkstra}. 
\item Training a surrogate model to predict correctness: this involves training a model or developing an empirical cost model to predict the correctness~\cite{mankowitz2023faster,10.1145/3630106.3658984}.
\item Limit the use of ML to small affine sequence of operations that could be validated by comparing the generated output to the canonical output~\cite{10.5555/3618408.3619263,cummins2023large}
\item Limit the use of ML to generating code based on pre-determined target-specific optimizations used in code of similar compute profile~\cite{10.1145/3577193.3593714}, or pre-determined optimizations in code manuals~\cite{DBLP:conf/iclr/Zhou0XJN23}.
\item Use symbolic execution~\cite{COTRONEO2024112113} or round-trip execution~\cite{DBLP:conf/icml/AllamanisPY24} to assess whether the AI-generated code behaves as a reference implementation.
\end{itemize}
In summary, the approaches above either provide no formal guarantee on functional correctness, or confine the user to a very constrained form of code generation. While some have suggested there is a \emph{tradeoff between correctness and performance},~\cite{ouyang2024kernelbench} we take a different view—emphasizing that correctness remains paramount. This perspective is reinforced by recent findings~\cite{2025sakana} where what was initially reported as a 100x speedup from an AI-generated solution was, upon closer examination, actually a 3x slowdown.   In addition, we observed that for the above approaches, it was the burden of the researchers to implement a sampling strategy and curate the dataset. Solving those two problems, correctness and agency, is the motivation for \TADASHI{}.



\subsection{The Polyhedral Model}
\label{sec:polyhedral}

Polyhedral techniques have a long history with one of the seminal papers dating back to 1996 \cite{feautrier1996efficient} and are well established and used in compilers~\cite{aho2007compilers}. The polyhedral model is a very efficient way to manipulate and reason about nested loops, including the capability to check the \emph{legality}\footnote{Legality is the technical term, used in the polyhedral literature with a precise definition which translates to the transformed code calculating the same results as the original, i.e., functional correctness.} of the transformations. It has been used in optimization pass of compilers~\cite{grosser2012polly}, as well as standalone source-to-source optimization tools~\cite{bondhugula2008practicala,verdoolaege2013polyhedral} relying on integer linear programming with some optimization heuristic.

\begin{lstlisting}[
  language=C,
  float,
  floatplacement=ht,
  basicstyle=\ttfamily\small,
  frame=lines,
  caption={
    A simple SCoP example with a single statement on line~\ref{line:stmt}, The statement has $N \times (M-1)$ instances, each identified by the iterator vector $(i, j)$, in general denoted as the integer vector $\vec{i}$.
  },
  label=lst:simple-scop]
for(int i = 0; i < N; i++)
  for(int j = 1; j < M; j++)
    A[i, j] += A[i, j-1]; /*@ \label{line:stmt} @*/
\end{lstlisting}

\begin{figure}
  \centering
  \subcaptionbox{$\theta(S[i, j]) = (i, j)$  \label{sf:schedule-i-j}}
  {\includegraphics{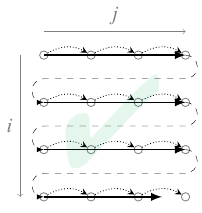}}
  \subcaptionbox{$\theta(S[i, j]) = (j, i)$  \label{sf:schedule-j-i}}
  {\includegraphics{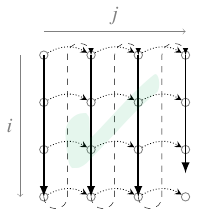}}
  \subcaptionbox{$\theta(S[i, j]) = (i+j, j)$\label{sf:schedule-ipj-j}}
  {\includegraphics{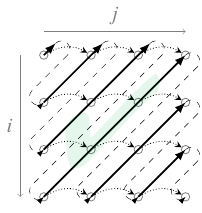}}
  \subcaptionbox{$\theta(S[i, j]) = (i, -j)$ \label{sf:schedule-i-mj}}
  {\includegraphics{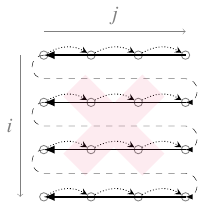}}
  \caption{
    Example schedules for Listing~\ref{lst:simple-scop}. The schedules are given in the sub-captions. The thick (and dashed) arrows show the execution order of the instructions.  The bent dotted arrows show the dependencies.  Schedule (\ref{sf:schedule-i-j}) is the original schedule, and all schedules other than (\ref{sf:schedule-i-mj}) are legal.
  }
  \Description{Example schedules}
  \label{fig:schedules}
\end{figure}

The polyhedral model is a representation for loop nests with regular boundaries and memory accesses, know as \emph{static control parts} (see Listing~\ref{lst:simple-scop} an example and Section~\ref{sec:quasiaffine} for a description). The polyhedral representation separates the statement instances from the order in which they are executed: Each of these instances is then mapped to an integer tuple, a timestamp, and then the statement instances are ordered according the lexicographical order of their timestamps. For a given statement $S$, the set of statement instances is called the \emph{domain} and is denoted as $\mathcal{D}_S$; while the mapping which determines their order is the \emph{schedule}, which is denoted by $\theta_S$. Listing~\ref{lst:simple-scop} and Figure~\ref{fig:schedules} give a simple example of a SCoP and possible schedules.


The main idea of polyhedral optimization is to transform the schedule (or create a new one from scratch) which reorders the statement instances so they are executed more efficiently on a given hardware (e.g., by increasing memory locality).
As one could expect, not all schedule transformations are legal, however checking legality is very convenient in the polyhedral model.
Dependency analysis~\cite{feautrier1991dataflow} yields a directed \emph{dependecy graph} $G$.
The vertices of $G$ are statements (or equivalently their domains), and the edges are labelled with maps which describe the dependencies on the level of statement instances such as
\begin{equation}\label{eq:dep}
S_i[\vec{d}] \mapsto S_j[\vec{r}]
\end{equation}
Applying the schedule $\theta$ to the dependee $S_i[\vec{d}]$ and dependent $S_j[\vec{r}]$ of these maps takes them into the time domain, and \eqref{eq:dep} becomes $t_d \mapsto t_r$ ($t_d$ and $t_r$ are timestamps).  If $t_d$ lexicographically precedes $t_r$, i.e.,
\begin{equation}
  \label{eq:correct}
  t_d <_{\text{LEX}} t_r
\end{equation}
then the schedule is legal, because it maps the dependent to a later time then the timestep of the dependee. Because \eqref{eq:correct} is equivalent to $0 <_{\text{LEX}} t_r - t_d$, we can (symbolically) compute $\delta=t_r - t_d$ and if $\delta$ is lexicographically positive, the schedule is legal. Computing $\delta$ is useful because it provides us with even more useful information: if an entry in $\delta$ is zero,
it means there are no dependencies along the corresponding dimension of the schedule, and consequently the generated loop for that dimension can be parallelized.

\subsection{Constraints of the Polyhedral Model}
\label{sec:quasiaffine}

To be able to represent the domain, schedule, and dependencies efficiently, the source codes represented by the polyhedral model are restricted to \emph{Static Control Parts}, or \emph{SCoP}s for short (Listing~\ref{lst:simple-scop} is a simple example). While these restrictions are not negligible, most of high performance and science codes do not violate them. The main restrictions of SCoPs are the following:
\begin{enumerate*}
  \item The control-flow must be static, which means it does not depend on the input, other than \emph{parameters} (such as loop boundaries) which remain constant during execution, 
  \item there is no memory aliasing, no pointer manipulation, and
  \item the symbolic expressions and formulas, used throughout the model, need to be expressed using a matrix $\mathbf{A}$ and a vector $\mathbf{b}$ as a (quasi-)affine expression, i.e., $\mathbf{A} \vec{i} + \mathbf{b}$ (sometimes referred to as Presburger formulas~\cite{verdoolaege2021presburger}).
\end{enumerate*}
For example, the domains are expressed as $\mathcal{D_S} = \bigl\{ S[\vec{i}] \in \mathbb{Z}^n : \mathbf{A} \vec{i} + \mathbf{b} \le \mathbf{0} \bigr\}$; schedules are (quasi-)affine functions $\theta(S[\vec{i}]) = \mathbf{A} \vec{i} + \mathbf{b}$.
The polyhedral model can also handle tiling, so the formulas include integer division with a constant as well, and this is what the ``quasi-'' prefix refers to. However, for brevity and clarity, we omit discussion of these details and for a more comprehensive description of the polyhedral model we recommend~\cite{feautrier1996efficient,feautrier1997efficient,verdoolaege2021presburger}.

Abiding by those restrictions, which is reasonably possible in a wide range of high performance science and engineering applications, enables ML systems that use \TADASHI{} to guarantee correct and legal output at a scale, unlike most other existing solutions.

\section{\TADASHI{}: A Library to Sample Legal Polyhedral Transformations}

\TADASHI{} is a library that provide the means for ML practitioners to sample the space of legal polyhedral loop transformations. The user can use \TADASHI{} to describe a sampling strategy that is suited for the ML approach, e.g., uniform sampling to curate datasets vs. guided exploration of the space of legal transformations. Figure~\ref{fig:teaser} gives an overview of \TADASHI{}. At an architectural level, \TADASHI{} consists of a Python front-end and a C/C++ back-end (aptly named \verb!ctadashi!).  The front-end provides a minimal and easy-to-use interface to sample from the space of possible transformations, check their legality, and generate the transformed code. More importantly, the front-end being in Python encourages fast adoption by ML practitioners since Python is the de facto language used in AI/ML frameworks. The back-end implements this functionality, using the Integer Set Library (ISL)~\cite{verdoolaege2010isl} and Polyhedral Extract Tool (PET)~\cite{verdoolaege2012polyhedral} polyhedral libraries.

Existing ISL wrapper libraries for Python expose the full API of the library, which can be daunting for users not versed in the concepts of polyhedral compilation.
The goal of \TADASHI{} is to lower this barrier to entry for ML engineers and researchers by providing out-of-the-box implementations of transformations, legality check and code generation.
Moreover, the functionality is exposed in a more idiomatically Pythonic way. For example, a tree of nested loops is presented is a hierarchy of nested objects similar to how HTML parse trees are used in Python libraries, i.e., a transformation on a given node can be done by calling an appropriate method of a given object, while hiding all bookkeeping necessary for interacting with ISL and PET libraries.

\subsection{Python Front-End: Polyhedral Made Easy}
To demonstrate the ease of use of \TADASHI{}, Listing~\ref{lst:end2end} performs a loop fission, checks legality, generates new code and measures speedup in no more than 10 lines of a fully functional Python example.  Line~\ref{ln:app} creates the \texttt{app} object, which stores the information required by \TADASHI{} to load the input file.  The location of the transformation is determined by selecting a \texttt{node} objects on line~\ref{ln:scops-nodes}.  The fission transformation is performed on line~\ref{ln:transform}.  The new code code is generated in line~\ref{ln:codegen}.
\begin{lstlisting}[
  float,
  floatplacement=H,
  frame=lines,
  basicstyle=\ttfamily\small,
  language=Python,
  caption={
    End-to-end demonstration of the use of \TADASHI{}, including transforming the code, checking the legality, generating the new code and measuring the speedup.
  },
  label={lst:end2end}]
from tadashi import TrEnum
from tadashi.app import Polybench
app = Polybench("linear-algebra/blas/gemm", ...)  /*@\label{ln:app}@*/
node = app.scops[0].schedule_tree[2] /*@\label{ln:scops-nodes}@*/
legal = node.transform(TrEnum.FULL_SPLIT) /*@\label{ln:transform}@*/
tapp = app.generate_code() /*@\label{ln:codegen}@*/
app.compile()
tapp.compile() /*@\label{ln:compile}@*/
speedup = app.measure() / tapp.measure() /*@\label{ln:measure}@*/
print(f"{legal=} {speedup=}")
\end{lstlisting}

To pass an input file to \TADASHI{}, it must be wrapped in an instance of the \texttt{App} abstract base class. Currently \TADASHI{} provides two implementations of \verb!App!: the \texttt{Simple} class which can be used for simple single-file programs and \texttt{ Polybench} (used in line~\ref{ln:app}) to load Polybench kernels~\cite{reisinger2024matthiasjreisinger}.  The classes derived from \verb!App! store the input filename to \TADASHI{} and other information needed by \TADASHI{} when parsing the file.  For convenience, \verb!App! also defines \verb!compile()! and \verb!measure()! methods.

The polyhedral representation of all loop nests (SCoPs) are implemented as an array-like \verb!scops! object (line~\ref{ln:scops-nodes}), and from the first SCoP (\verb!scops[0]!) a \verb!node! of the \emph{schedule trees}~\cite{verdoolaege2014schedule} is selected.  For a user not familiar with the polyhedral model and schedule trees, a \verb!node! is ``an entity that can be transformed'', which usually means a loop in the loop nest (a brief discussion of schedule trees can be found in Section~\ref{sec:under-the-hood}).  Since some transformations are not available for certain loops (e.g., it doesn't make sense to apply 3D tiling on a 2-deep loop nest), the \texttt{Node} class provides methods to query the available transformations and their (valid) arguments. The \verb!node!s of a \verb!scop! can be accessed through the (dynamic) \verb!scop.schedule_tree[]! as a list, or as a tree using the \verb!goto_child(child_idx)! and \verb!goto_parent()! methods of \verb!node!s.\footnote{The root is \verb!scop.schedule_tree[0]!.}

A code transformation is performed by passing a \texttt{TrEnum} (an \texttt{Enum} type enumerating all transformations) object to the \texttt{node}'s \texttt{transform()} method (with the transformation-specific argument).  A summary of implemented transformations (and their parameters) can be found in Table~\ref{tab:transformations}. \texttt{TILE1D}, (\texttt{TILE2D} and \texttt{TILE3D}) implements 1D, (2D and 3D) tiling (higher dimentional tiling can be expressed tiling and \texttt{INTERCHANGE} transfromations). \texttt{INTERCHANGE} switches an outer loop with the loop directly under it. For \texttt{PARTIAL} shift transformations, the statement index (stmt idx) selects only one statement to be shifted, while the \texttt{FULL} versions shift all statements.  The \texttt{VAL} shift transformation shift the schedule by a constant, the \texttt{VAR} and \texttt{PARAM} versions shift by an iterator (iter idx) of an outer loop or a parameter (param idx),  multiplied by a coefficient (coeff). \texttt{FULL\_FUSE} fuses all loops in a sequence, while \texttt{FUSE} fuses two selected loops. Similarly, \texttt{FULL\_SPLIT} and \texttt{SPLIT} perform loop fission.  \texttt{SET\_PARALLEL} and \texttt{SET\_LOOP\_OPT} modify how code of a loop is generated. These are primitive transformations, in the sense that they are \emph{composable}, so we can apply a sequence of transformations on the same SCoP. In the current version of \TADASHI{}, we prioritized the the implementation of loop transformations that have high impact on performance~\cite{10.1145/1926385.1926449}.

The return value of the \texttt{transform()} method is the result of the legality check and, based on it, the users can decide what they want to do.  In case of a legal transformation, one likely action would be generating the C code (line~\ref{ln:codegen} in Listing~\ref{lst:end2end}), compiling it (line~\ref{ln:compile}) and measuring the execution time (line~\ref{ln:measure}), while for an illegal transformation one could invoke the \texttt{rollback()} method of the node, to return the polyhedral representation to the previous, legal state.  Additionally, \texttt{Scop} objects provide a reset method and a method to apply a list of transformations, useful to implement ``arbitrary rollback''.

\begin{table}[t]
  \caption{
  Transformations (primitives) implemented in \TADASHI{}.}
  \centering
  \begin{tabular}[h]{ll}
    \toprule
    \verb!TrEnum! & Args \\
    \midrule
    \verb!TILE! & tile size \\
    \verb!INTERCHANGE! & -- \\
    \verb!FUSE! & 2 loop indices \\
    \verb!FULL_FUSE! & -- \\
    \verb!SPLIT! & 1 stamt index \\
    \verb!FULL_SPLIT! & -- \\
    \verb!FULL_SHIFT_VAL! & const shift value \\
    \verb!PARTIAL_SHIFT_VAL! & stmt idx, const shift value \\
    \verb!FULL_SHIFT_VAR! & coeff, shift iter idx \\
    \verb!PARTIAL_SHIFT_VAR! & stmt idx, coeff, shift iter idx \\
    \verb!FULL_SHIFT_PARAM! & coeff, shift param idx \\
    \verb!PARTIAL_SHIFT_PARAM! & stmt idx, coeff, shift param idx \\
    \verb!SCALE! & coeff \\
    \verb!SET_PARALLEL! & -- \\
    \verb!SET_LOOP_OPT! & \verb!AstLoopType! enum \\
    \bottomrule
  \end{tabular}

  \label{tab:transformations}
\end{table}

\subsubsection{Sampling Strategies}
\label{sec:strategies}
Listing~\ref{lst:end2end} demonstrates the application of a single primitive transformation.  However, in consequence of the compositionality of polyhedral transformations, the user can apply a sequence of transformations and legality checks to a SCoP (before generating the code) and hence obtain a different \emph{composite} transformation.  This is a \emph{new} sample from the space of polyhedral transformations, different from its component primitive transformations.  Put differently, sampling the space of polyhedral transformations with \TADASHI{} consists of applying a sequence of transformations to a SCoP, where each entry consists of specifying a location (a node in the schedule tree), the transformation (\verb!TrEnum!) and a set of arguments.  A location can be selected using the \verb!schedule_tree[]! member or with a combination of \verb!goto_*! methods of the \verb!Node! class.  The transformation can be selected from the possible values \verb!TrEnum! values and checked if it is available on the selected node.  Finally, selecting an argument is somewhat convoluted, since the available values depend not only on the selected transformation, but also on the concrete properties of the selected schedule node (such as its partial schedule, type and number of children etc.).  The \texttt{get\_args()} method of \texttt{Node}s simplifies this.

\subsection{Using Different ML Methods With \TADASHI{}}
The goal of \TADASHI{} is to lower the barrier to entry for user-guided sampling of legal loop transformations. \TADASHI{} does not in itself implement any type of ML; we do however design \TADASHI{} with considerations to how various ML approaches can use it. The design of \TADASHI{} supports the following ML search space exploration approaches:

\subsubsection{Reinforcement Learning (RL)}
RL~\cite{sutton2018reinforcement} is a prime candidate to use \TADASHI{}, since \TADASHI{}'s interface maps naturally to RL concepts.
Similar to choosing winning moves in a game, an RL agent chooses promising transformations.
The transformed code is executed by the runtime and speedup over the original version serves a the reward signal. Sequences of transformations can be sampled by using approaches such as Monte Carlo tree search (MCTS), where a neural network model gives scores to possible choices - an approach similar to Alpha Go\cite{Silver2016} system.


\subsubsection{Supervised Learning}
Different versions of transformed source codes and corresponding speedups can be saved in a database.
A model to predict speedups from transformations 
can be trained in supervised fashion to be used as a value model in an off-policy reinforcement learning setup or in other applications. 

Such approach is especially valuable, because evaluation of transformed code takes significantly more time, compared to selecting and performing transformations (see Section~\ref{sec:overhead}).

The choice of transformations can be done in random or by grid search. 
Many transformations require additional corresponding parameters, e.g., choosing to tile a loop requires choosing a tile size. 
These arguments (second column in Table~\ref{tab:transformations})
allow the user to combinatorially increase the coverage of the search space to consequently increase the coverage of the curated dataset. For example, starting with a limited set of input C source files containing loop nests, one could run \TADASHI{} to select random transformations (on a random input file, on random nodes of the schedule tree, with random parameters), then store the new code with the result of the legality check and the measured execution time.  The dataset generated this way would consist of (input, output) pairs, where the input is (the name of) a C source file, and the output consists of:
\begin{enumerate*} 
\item the transformation, 
\item transformed code, 
\item legality of the transformation, 
\item execution time of the transformed program.
\end{enumerate*}

In the case of legal transformations, the transformed code can be added to the pool of input files, expanding the number of different input files \emph{ad infinitum}.

Another, approach is to sample sequences of transformations, i.e., generate a sequence (of a random length) of random transformations, where the individual transformations are sampled similarly as described in the previous paragraph.  The legality check, code generation and performance measurement are performed only at the end of the sequence.

With basic Python programming, e.g., using the \verb!itertools! package, one can easily generate a regular ``grid'' of transformations, which is adequate for scenarios similar to auto-tuning.

\subsubsection{Evolutionary Computation Algorithms}  Similar to RL, evolutionary computation (EC) algorithms use a mix of exploration and exploitation to traverse the search space, however without relying on learnable parameters and performing search using more direct heuristics~\cite{CAI2023119656}. Exploration refers to searching for new, diverse solutions by exploring a wide range of the solution space, while exploitation focuses on refining and improving existing solutions by intensively searching around the best solutions found so far. As mentioned above in other ML approaches, \TADASHI{}'s sampling of the search space can be parameterized by the user to explore a wide range of the search space (exploration) or intensive search in a narrow region of the search space (exploitation). The runtime of the transformed code variants would be used as the objective function.

The exploration capability of EC algorithms can in some instances improve if infeasible solutions appear in the population~\cite{6557723}, and then later be penalized to disappear over time from the population~\cite{10.1145/3520304.3533640}. \TADASHI{} internally manipulates the schedule to generate both legal and illegal transformations, where we then filter out the illegal transformations. Since illegal transformations can be of value for the exploration capability of EC algorithms, we allow the user to sample illegal transformations simply by not rolling them back and generating code disregarding the result of the legality check.

\subsubsection{With LLM Agents} 
The most performant transformations identified by any of the aforementioned methods can be used  as training data for Large Language Models for better end-to-end code generation or as additional signal to improve overall capacity of the model~\cite{10.1145/3597503.3639219}.

However, in line with the motivation for this paper, an even more promising application of \TADASHI{} would be to be called directly by an LLM agent in a ``tool calling'' paradigm \cite{ToolFormer}, similar to Retrieval Augmented Generation (RAG) systems~\cite{lewis2020retrievalaugmented} or Chain of Thought approach~\cite{Wei_CoT}.


\subsubsection{Auto-Tuning}
For practical reasons, ML approaches used in auto-tuning~\cite{9698048,10.1145/3197978} mostly rely on local search methods with focused search around particular regions in the search space, after aggressively pruning the search space. Similar to the exploitation in EC algorithms, \TADASHI{} can be setup to narrow down the search around the region for auto-tuning, based on the post-pruning range of interest.

\subsubsection{ML-driven deployment of applications on new hardware} In addition to supporting the usage patterns of the ML approaches listed above, \TADASHI{} can drive an ML approach to learnhardware specific optimisations on new hardware. After the training phase, applications ported to the new hardware can do inference on the ML model, leveraging \TADASHI{}, to optimize the applications for the new hardware.

\section{Implementation of \TADASHI{}}
\label{sec:under-the-hood}

\begin{figure}[t]
  \centering
  \includegraphics[width=\linewidth]{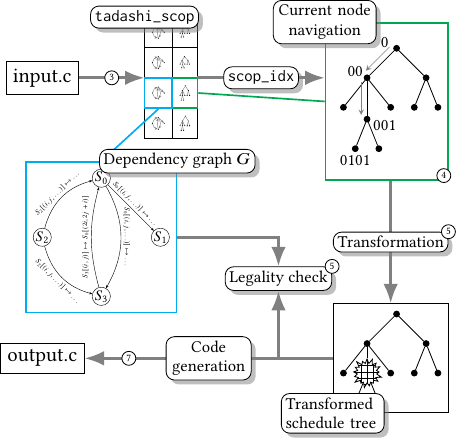}
  \caption{Visual representation of the \texttt{ctadashi} back-end.  Labels in the small circles refer to the line numbers in Listing~\ref{lst:end2end}.}
  \Description{Overview of the ctadashi back-end.}
  \label{fig:workflow}
\end{figure}

We now revisit the end-to-end example of Listing~\ref{lst:end2end}, but with our attention directed to implementation details and what is happening under the hood, in the \verb!ctadashi! back-end depicted in Figure~\ref{fig:workflow}.

\subsection{The App Class}
The \verb!App! class (Listing~\ref{lst:end2end} line~\ref{ln:app}) is completely written in Python and provided mostly for convenience. It's main functions are to provide the \verb!source_path! and \verb!include_paths! properties, the former being the C source file which will be transformed, and the later a list of paths used to set the required environment variable,\footnote{Namely \verb!C_INCLUDE_PATH!.} which is necessary, because when PET runs a compiler pass on the input file, it will throw an error if it cannot find an included header file.

\subsection{SCoPs}
Accessing the \texttt{Scops} object (line~\ref{ln:scops-nodes}) sets things in motion in the C/C++ back-end of \TADASHI{}.  The input file is passed through PET's C source transformation function\footnote{Namely \texttt{pet\_transform\_C\_source}.} which detects all the SCoPs in the file, but instead of doing any transformations, the \texttt{pet\_scop} objects are used to populate a container \texttt{tadashi\_scop} structs.  These structs store bookkeeping information and ISL objects, such as the domain, schedule and dependencies, required for the legality checks. Dependencies are computed by ISL as described in~\cite{verdoolaege2021presburger} and algorithm adopted from PPCG~\cite{verdoolaege2013polyhedral} is executed at this point to eliminate dead code. \TADASHI{} also handles the management of loading multiple apps, requiring the creation, storage and deletion of multiple \texttt{Scops}.

\subsection{Schedule Tree and Navigation}
Transformations in \TADASHI{} are performed on schedule trees, which are alternative ways to represent the domains and schedules in a SCoP (see Section~\ref{sec:polyhedral}). The leaves of the schedule tree are the statements, the root is the union of their domains. The loops surrounding a statement (in the source code) correspond to \emph{band nodes} in the tree. On the path from the root to the leaf, each band node has a partial schedule and the complete schedule of the statement (i.e., leaf) can be constructed by concatenating the partial schedules. Other nodes in the schedule tree are sequence and filter nodes. Each \emph{sequence node} has multiple \emph{filter nodes} as children and together they represent a sequence of statements and/or nested loops. The sequence node is the placeholder node for the entire sequence, while its children act as roots, i.e., domains for the sub-schedule trees.

Each transformation in \TADASHI{} is defined on a node of a schedule tree, so line~\ref{ln:scops-nodes} in Listing~\ref{lst:end2end} involves the following mechanism.  First, \texttt{scops[i]} returns a Python \texttt{Scop} object which stores the index of the corresponding \texttt{tadashi\_scop} entry on the C++ side. This \texttt{Scop} object wraps all the C++ functions exposed to Python by calling them with a \texttt{scop\_idx} parameter from an object method with same name.  These functions include navigation operations\footnote{Namely \texttt{goto\_root()}, \texttt{goto\_parent()}, and \texttt{goto\_child(idx)}.} to update the current node to point to its parent, its child, or to the root of the schedule tree (for the selected \texttt{tadashi\_scop} object).  The \texttt{schedule\_tree} property on line~\ref{ln:scops-nodes} is dynamically generated, and exposes the schedule tree as a list of nodes by doing a depth-first traversal.  During the traversal, we build a list of child indexes which enables \TADASHI{} to locate any node by calling \texttt{goto\_child} for each element in the list (implemented as the \texttt{locate()} method of a \verb!Node!).  In addition to the navigation methods, \TADASHI{} exposes wrappers to query properties of nodes, (such as node type, number of children, partial schedule, parameters, enclosing iterators etc.), and so enables the Python \texttt{Node} object to emulate the ISL schedule tree and its nodes in C++.

\subsection{Transformations}
After updating the current node to point to the desired location, invoking \verb!node.transform()! calls the C++ function implementing the corresponding transformation.  The transformation is performed on the current node of the SCoP specified by the index obtained from the \verb!node! (similar to the navigation and query functions from the previous paragraph).  To be able to rollback the transformation, the current node is copied, and restored if the user calls the \verb!rollback! method (usually after the transformation turns out to be illegal).

\subsubsection{Implementation Details of Individual Transformations}
Multi dimensional tiling is already implemented in ISL, and to expose a simple 1D tiling (with a single integer as its argument) we convert the tile size into the a single valued ISL \verb!isl_multi_val! object.

Loop interchange is can easily be implemented by: 
\begin{enumerate*}
  \item saving the partial schedule of the current node,
  \item deleting the current node, and
  \item inserting a new node with the saved schedule below the remaining node.
\end{enumerate*}

Fusing of two loops in a sequence (\verb!FUSE!) is achieved by collecting the two filters (the children of the original sequence node) in an \verb!isl_union_set_list!, then replacing the filters
of the two nodes by a single (union) filter, and finally inserting a new sequence node on top of the original sequence node.

To merge all band node under a sequence (\verb!FULL_FUSE!), we take their partial schedules, intersect them with the corresponding filters and take the union. Then we introduce a new band node on top of the sequence\footnote{Using \
\verb!isl_schedule_insert_partial_schedule!.} and  delete the original band nodes.

The 6 different \verb|SHIFT| transformations, while simple on the surface, involve a tedious process involving the unpacking of the partial schedule, iterating through the sets of its domain and (re)constructing the adequate \verb!mupa!\footnote{Short for \verb!isl_multi_union_pw_aff!.} object.

The \verb!SET_PARALLEL! transformation simply insert a ``mark'' node, with an \verb|isl_id| name \verb|"parallel"| and the code generatation handles the rest.

Similarly to \verb|SET_PARALLEL|, \verb!SET_LOOP_OPT! is not strictly speaking a (polyhedral) transformation. It is implemented by simply calling the corresponding ISL function, and influences how code ISL generates code.

\subsection{Legality Checking}
\label{sec:legal}
\begin{figure}[t]
  \centering
  \includegraphics[width=\linewidth]{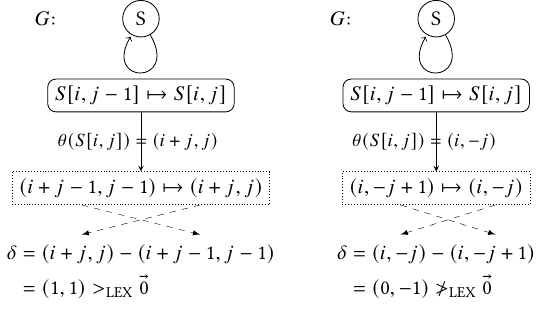}
  \caption{Example of legality checking.
  }
  \Description{Flow of legality check for two examples.}
  \label{fig:legality}
\end{figure}
After a transformation concludes, the legality check is invoked on the same node. The legality check implements the principles described in Section~\ref{sec:polyhedral}.  Each step of the legality check in Algorithm~\ref{alg:legality} can be implemented with a single ISL function.  Algorithm~\ref{alg:legality} is applied after every transformation with the exception of \verb!SET_PARALLEL!.  For the legality of parallel loops the entry in $\delta$ corresponding to the loop should be $0$, and this can be achieved with a slight modification to Algorithm~\ref{alg:legality}: we use the \emph{partial} schedule instead of the complete schedule in steps~\ref{aln:dmn}~and~\ref{aln:rng}, and in step~\ref{aln:leset} we compute $M_{<}=\{ \vec{d} \to \vec{0} : \vec{d} \in \delta \land \vec{d} <_{\text{LEX}}\vec{0} \}$ and $M_{>}=\{ \vec{d} \to \vec{0} : \vec{d} \in \delta \land \vec{d} >_{\text{LEX}}\vec{0} \}$ and check if they are both empty in step~\ref{aln:check}. Figure~\ref{fig:legality} shows two examples of the process of checking legality. Checking legality starts with the dependency graph $G$ at the top, which, in case of Listing~\ref{lst:simple-scop}, is one vertex with a loop edge.
    The domain and range of the map (below $G$), which indicates that statement instance $S[i, j]$ depends on $S[i, j-1]$, are transformed according to the schedule $\theta$.
    The resulting transformed map is shown below (dotted rectangle).
    If the difference between the (transformed) domain and  the (transformed) range, denoted by $\delta$, is lexicographicaly positive, the schedule is legal (left side, the $\theta$ from Figure~\ref{sf:schedule-ipj-j}), otherwise the schedule is not legal (right side, the $\theta$ from Figure~\ref{sf:schedule-i-mj}).
\begin{algorithm}
  \caption{
    Legality check (The modifications used for \texttt{SET\_PARALLEL} are described in the main text.)
  }\label{alg:legality}
\begin{algorithmic}[1]
  \REQUIRE The dependencies ($G$) and the schedule associated ($\theta$)
  \STATE Apply $\theta$ to the domain of the dependencies map \label{aln:dmn}
  \STATE Apply $\theta$ to the range of the dependencies map \label{aln:rng}
  \STATE Calculate their $\delta$ (i.e., the difference of the two previous results)
  \STATE Compute the map $M_{\le}=\{ \vec{d} \to \vec{0} : \vec{d} \in \delta \land \vec{d} \le_{\text{LEX}}\vec{0} \}$ \label{aln:leset}
  \RETURN {Legal if $M_{\le}$ = $\emptyset$, illegal otherwise} \label{aln:check}
\end{algorithmic}
\end{algorithm}

\subsection{Code Generation}
Code generation is based on the \verb|pet_loopback.c| file from the PET repository, which extracts the polyhedral representation from an input C file, and generates the ``same'' code using the original schedule.
Apart from the obvious tweaking to use the modified schedule instead of the original one, and to make it cooperate with the bookkeeping setup we use in \TADASHI{}'s back-end, the \texttt{SET\_PARALLEL} transformation required additional modifications to the core code generation. 

\section{Evaluation}
\label{sec:eval}
We evaluate the capabilities of \TADASHI{} by verifying functional correctnes, using two different ML solutions on top of \TADASHI{} to showcase how \TADASHI{} enables ML solutions to speedup different applications and benchmark suites.  Then we demonstrate how \TADASHI{} can be integrated with existing scientific computing codes, illustrating \TADASHI{}’s potential to enabling AI-based automated code generation with guaranteed correctnesss for scientific applications, and finally measure the overhead.

\subsection{Empirical Functional Testing}
\label{sec:randomrl}
We provided correctness guarantees of the polyhedral method we developed for \TADASHI{} in Section~\ref{sec:legal}. However, we discuss in this section how we empirically do functional testing to assure the correct implementation of \TADASHI{}. We evaluated \TADASHI{} by applying a range of primitive (single) and composite (combined) transformations on kernels from the Polybench~\cite{reisinger2024matthiasjreisinger} benchmark suite. 

To verify the correctness of implementation of \TADASHI{}, 
first, all Polybench benchmarks were compiled and executed with the small dataset\footnote{Using \verb!-DSMALL_DATASET!.} and with the output saved.\footnote{Using \verb!-DPOLYBENCH_DUMP_ARRAYS!.} In an automated manner, we generate 3 random transformations,
rollback the illegal ones, compile, run, and compare the output of the transformed program to the original. 
Post-inspecting confirmed the outputs to be identical. 

\subsection{Using \TADASHI{} with ML Solutions}
\label{sec:ml-apps}
\begin{figure}
    \centering
    \includegraphics[width=\linewidth, trim={10 12 10 10}, clip]{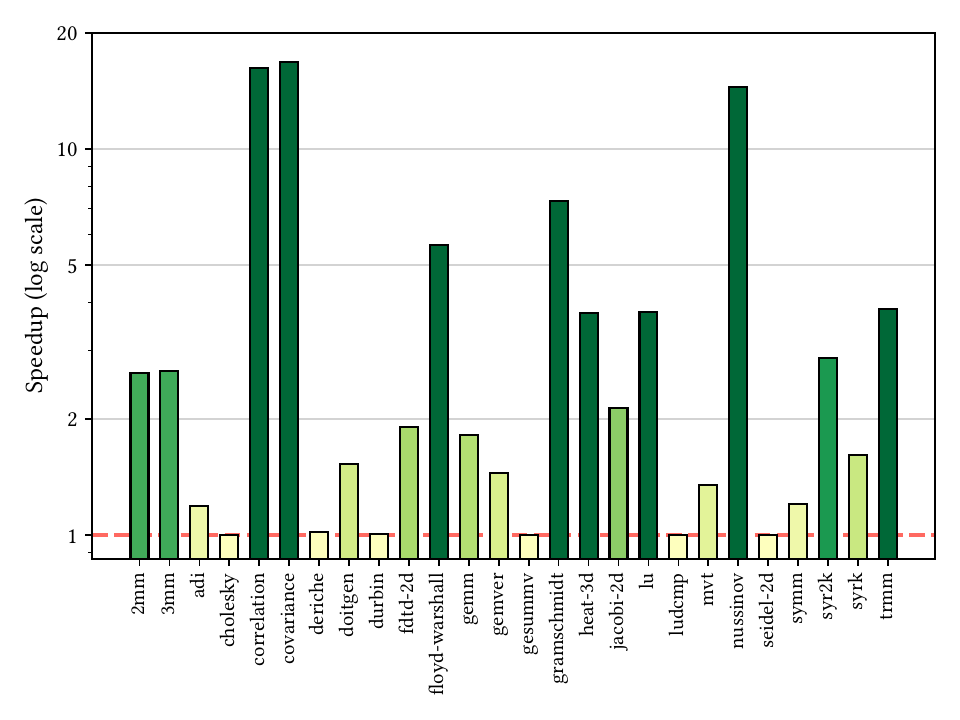}
    \caption{Speedup of Polybench suite~\cite{reisinger2024matthiasjreisinger} using POC-search.}
    \Description{POC algorithm speedups}
    \label{fig:poc_improvements}
\end{figure}

To demonstrate the practical utility of \TADASHI{} in real-world ML workflows, we developed and deployed three illustrative examples using our framework. The first is a simple proof-of-concept implementation of an optimal transformation search (POC-search), designed to highlight \TADASHI{}’s ability to apply, verify and evaluate transformations: While experimenting with transformations, we noticed fission and tiling (2D and 3D) can often speed up Polybench kernels,\footnote{Using the extra large dataset.} so with a fixed tile size, we implemented a brute-force search, applying fission and tiling when ever possible, rolling back and discarding illegal results to obtain speedups presented in Figure~\ref{fig:poc_improvements}. The \emph{atax}, \emph{bicg}, \emph{gesummv}, and \emph{jacobi-1d} kernels are not included since their execution time is under 0.01s and, as such, improvements on the results can be caused by noise in the CPU. A speedup of 1x indicates that no valid operations were found (\emph{cholesky}, \emph{deriche}, \emph{durbin}, \emph{ludcmp}, \emph{seidel-2d}) or that the transformation did not result in a speedup (\emph{gesummv}).

The other two are representative ML models that showcase how \TADASHI{} enables verifiably correct code generation in the context of ML. Each of these examples illustrates different aspects of \TADASHI{}’s design and capabilities. 

\subsubsection{Monte-Carlo Tree Search}
We have adopted Monte Carlo Tree Search (MCTS)~\cite{MCTS_Coulom} as a baseline method for discovering effective transformations and compositions of transformations.
MCTS algorithm iteratively builds a search tree where nodes represent states of the code after applying specific transformations (such as tiling, fusion, or interchange), and edges represent the transformations themselves. 
More precisely, we break each transformation down into three steps: selecting which node to transform, which transformation to apply and which parameters (e.g., tile size) to select for a given transformation.
This approach naturally handles the composition of transformations: applicability and effect of each transformation depends on previously applied ones. 
MCTS requires  no domain-specific heuristics beyond a performance evaluation function, thus allowing it to discover non-intuitive transformations.

Furthermore we employ the Upper Confidence Bound for Trees (UCT)~\cite{UCT_Kocsis} strategy to balance exploration of novel transformation sequences with exploitation of already discovered effective patterns.

\begin{figure}
    \centering
    \includegraphics[width=\linewidth, trim={0 9 0 40}, clip]{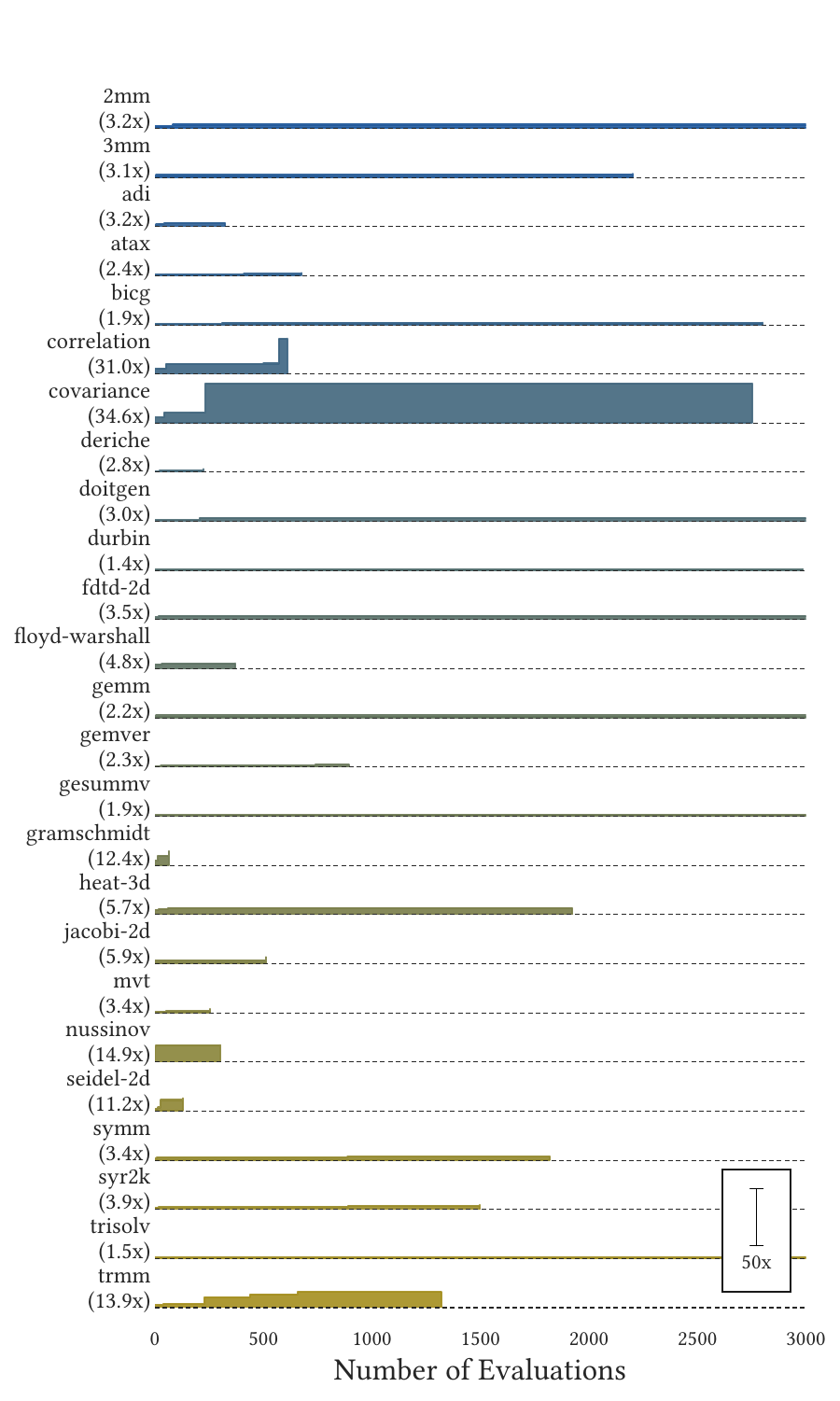}
    \caption{Example of a pace of discovering good chains of transformations during a single session of MCTS optimization on a range of Polybench kernels.}
    \Description{Plot of speedups discovered by MCTS on different benchmarks}
    \label{fig:mcts-ridge}
\end{figure}

Figure \ref{fig:mcts-ridge} illustrates the pace of discovery of effective transformations for a range of kernels from the Polybench suite. We excluded the loop parallelization transformation since in most cases it is easy to do manually, thus we wanted to focus on single-core performance improvement.
Each chart is a single run of \TADASHI{} on an AMD EPYC 9684X 96-Core Processor.
The main cost of running \TADASHI{}  is the actual execution of the transformed code. 
We aimed at execution time of about 10 seconds for a given kernel, to strike a balance between reliability of measurements and reasonable turn-around time. Since the execution time can be set arbitrary, we report performance improvement over the number of evaluation steps.

In line with our intuition for the POC approach, MCTS quickly discovered that as stand-alone transformations, interchange (e.g., for \emph{gemm}) and tiling (e.g., for \emph{floyd-warshall}) are often very effective. In terms of compositions, we see interchange or full split combined with tiling being discovered on multiple occasions. 
As a concrete example, transformations resulting in 6x performance improvement in \emph{trmm} kernel were discovered as follows:
\begin{verbatim}
trs = [
  [3, "FULL_SPLIT"],
  [1, "TILE1D", 3],
  [5, "INTERCHANGE"]
]
\end{verbatim}
The first number is the index of a node in the polyhedral schedule tree, last number indicates the transformation parameter.
More examples can be found in the \texttt{examples/evaluation/data} directory in the project repository.


\subsubsection{Evolutionary Computation}
In addition to the MCTS, we explored using a Genetic Programming~(GP)-based~\cite{Vanneschi2023} approach to search for a good set of transformations. Unlike MCTS, GP is allowed to use all transformations provided by \TADASHI{}. We limit the number of iterations to 20 and use a population size of 20, i.e., the algorithm does a total of 400 evaluations to obtain the final list of operations. 

In this implementation, a model is represented as a list of operations to be applied to a SCoP. The models are initialized as an empty list of transformations. During training, the models can be mutated to append a randomly selected transformation to their list or lose one operation, and can also be combined with other models by picking a cutting point from both models' lists and swapping the latter halves, creating two new models. Any invalid models generated this way are discarded.

The GP experiments are still in a preliminary stage, and while we only obtained results for two of the Polybench benchmarks (\emph{gemm} and \emph{jacobi-2d}), the results are promising. On \emph{gemm}, GP obtained a 18.2x speedup (4.0x, without using the parallel transformation) after 12 iterations (240 evaluations). On \emph{jacobi-2d}, GP obtained a 2.8x speedup after 8 iterations (160 evaluations). These results indicate that GP can beat the results previously seen in Figure~\ref{fig:poc_improvements}. 

\subsubsection{Transformations Found by \TADASHI{}}
The following lists show two examples of the \TADASHI{} transformations that GP found to speedup \emph{gemm} and \emph{jacobi-2d} by 18.2x and 2.8x, respectively. Additional details of transformations that \TADASHI{} discover are included in the supplementary material, as permitted for accepted papers. Each row contains the \texttt{node} index, the transformation and its arguments.
\begin{verbatim}
gemm = [ 
  [1, "FULL_SHIFT_PARAM", 2, 16], 
  [2, "FULL_FUSE", ], 
  [1, "SET_PARALLEL", 6], 
  [2, "FULL_SHIFT_PARAM", 1, 48], 
  [2, "TILE2D", 32, 32], 
  [3, "SET_LOOP_OPT", 0, 3],
]
jacobi-2d = [ 
  [9,  "TILE1D", 32], 
  [4,  "TILE2D", 32, 32], 
  [10, "TILE2D", 32, 32], 
  [10, "TILE1D", 32], 
  [13, "INTERCHANGE"],
]
\end{verbatim}

\subsection{Integration with Scientific Applications}
\label{sec:proxyapps}
To evaluate the applicability of \TADASHI{} into more realistic scientific computing scenarios, beyond simple benchmarks overspecialized for polyhedral optimization, we integrated it with two representative applications: \emph{miniAMR}~\cite{vaughan2015enabling} and \emph{FDTD electromagnetic simulation} from~\cite{sullivan2013electromagnetic}. These codes, while manageable in size, reflect the complexity and structure of production-level scientific software, providing valuable insight into how \TADASHI{} can support correctness and agency in practical settings.

\emph{miniAMR} is a proxy application developed as part of the Mantevo project, designed to emulate the behavior of Adaptive Mesh Refinement (AMR) codes used in large-scale physics simulations. It features block-structured grids, dynamic refinement and coarsening, and stencil-based updates—characteristics commonly found in production AMR codes.

The \emph{FDTD electromagnetic simulation} uses the Finite-Difference Time-Domain (FDTD) method. It implements a full 3D simulation kernel including absorbing boundary conditions, and serves as a self-contained, practical example of a numerical method widely used in computational electromagnetics.

The \emph{FDTD electromagnetic simulation} is relatively simple and self-contained, with a clear computational kernel, which made it well suited for \TADASHI{}-driven optimization--resulting in a 13x speedup. \emph{miniAMR}, on the other hand, is significantly more complex, featuring dynamic mesh refinement and distributed data structures. Despite these challenges, \TADASHI{} was able to produce a 1.05x (5\%) speedup, demonstrating utility even in more intricate scientific codes.

The modest 1.05x speedup in \emph{miniAMR} highlights the need for more sophisticated machine learning techniques and a more elaborate setup to fully exploit the potential of \TADASHI{} in complex applications. Specifically, a more refined strategy for distributing the evaluations and measuring the runtime across multiple nodes would likely yield better scaling of ML methods. Fortunately, \TADASHI{} fully supports such a distributed setup, leveraging tools like \texttt{MPI4Py.futures} to enable parallel execution and scalable optimizations across a cluster. This capability will be crucial as we continue to refine our approach and tackle even more complex scientific workloads.

\subsection{Overhead} 
\label{sec:overhead}

\begin{figure}[t]
  \centering
  \includegraphics[width=\linewidth, trim={0 12 0 0}, clip]{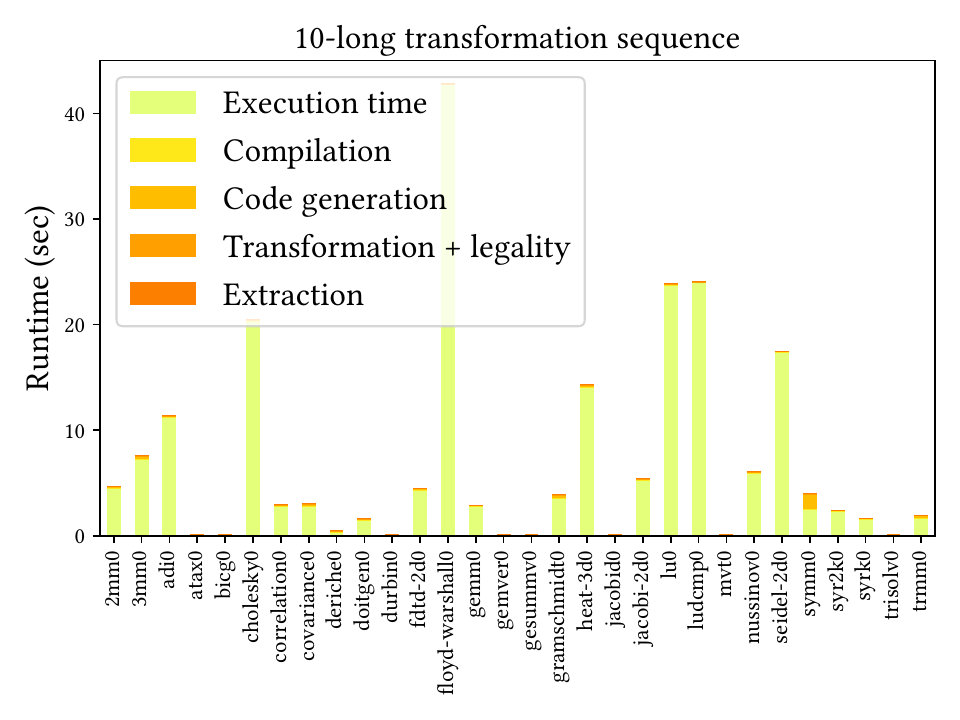}
  \caption{
  Breakdown of execution times of \TADASHI{}'s functions for 10-long sequence of transformations.
  }
  \Description{Breakdown of the results}
  \label{fig:breakdown}
\end{figure}

\begin{figure}[t]
  \centering
  \includegraphics[width=\linewidth, trim={0 12 0 0}, clip]{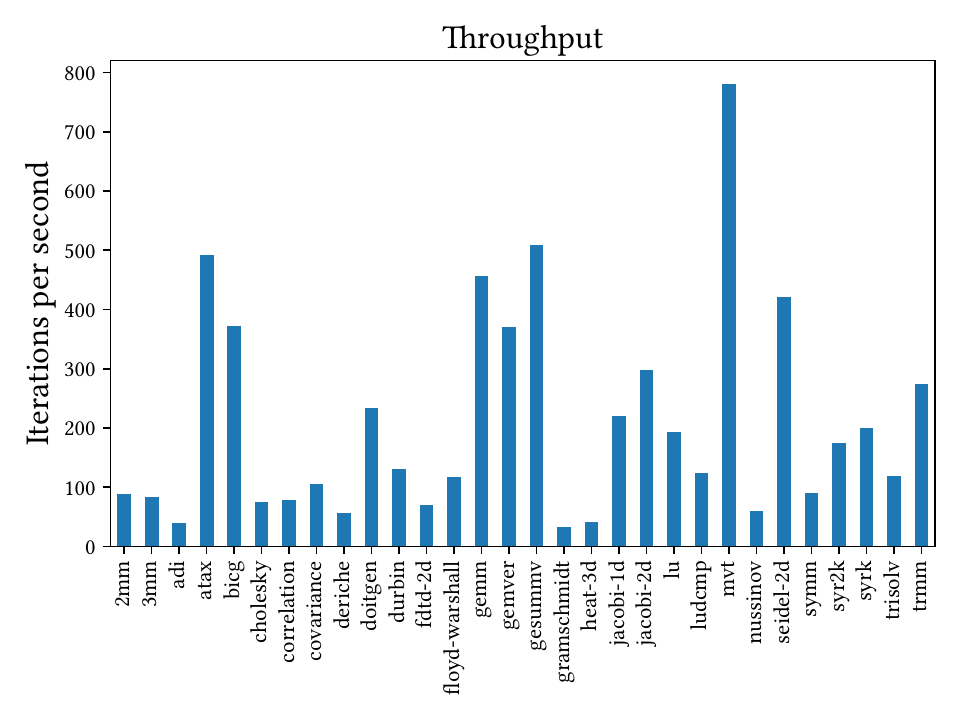}
  \caption{
  Throughput of transformations (and legality checks) for all Polybench apps.  Each iteration consists of a SCoP transformation and a legality check.
  }
  \Description{Throughput}
  \label{fig:throughput}
\end{figure}

ML training algorithms need a vast number of code transformations, including the legality check and code generation, it would be informative to see how fast \TADASHI{} performs these functions.  With this in mind, we devise two experiments to measure a typical training cycle: one which involves a single polyhedral transformation and another with a sequence of 10 transformations (and legality checks), with a code generation at the end. A typical training cycle would cover functions performed by \TADASHI{}, such as, extracting the polyhedral representation of the SCoPs, performing the transformation, checking the legality, generating code, as well as, compiling and running the program; steps with execution times which neither the user nor \TADASHI{} can influence or circumvent. For both experiments, we use Polybench with the default dataset size, based on the reasoning that a smaller dataset would fit into cache memory and shrink the runtime to an extent where optimization would not be required.

The breakdown of the results can be found in Figure~\ref{fig:breakdown}.  The dominance of the runtime of the programs is obvious in more then half of the apps. As for the other benchmarks, with miniscule runtimes, we found that the functions performed by \TADASHI{} always occupied less then half of the breakdown. Based on these numbers, using \TADASHI{} should not be considerably more prohibitive than any other workflow which involves compiling and/or running the program.

The overhead measurements are done on a regular desktop with a 12 core AMD Ryzen 5 5600X 6-Core Processor and 64GiB of memory, running Arch Linux, with python 3.12.5-1, libisl 0.27-1, llvm 18.1.8-4 system packages, and \TADASHI{}'s CMake script using the current master branch of PET\footnote{From \url{https://repo.or.cz/pet.git}.} tagged as pet 0.11.8.

To provide a better insight into \TADASHI{}'s performance, we also calculate throughput from the breakdown data,  which is shown in Figure~\ref{fig:throughput}.  Only the number of transformations and legality checks are counted, since it can happen more often than SCoP extraction and code generation, and the latter two would also be accompanied by compilation and program execution, which, as we saw in Figure~\ref{fig:breakdown}, would dominate the runtime. The results show \TADASHI{} capability of sustaining 100s of iterations per second, which would be very practical for production ML solutions where 100,000s $\sim$ 100,000,000s of training examples or direct evaluations in RL would be the norm.

\section{Conclusion}
In this work we aim to lower the barrier for using ML in code auto generation, with guarantees on correctness. We rely on the polyhedral methods to generate code based on compositional loop transformations with mathematically proven guarantee of functional correctness.
To this end we introduce \TADASHI{}, a library which exposes an API simplified to the polyhedral methods, targeting the ML engineers and researchers not familiar with it. By using \TADASHI{}, ML practitioners can design ML-driven methods based on the capability of \TADASHI{} in sampling of the space of legal loop transformations. 

While our initial experiments with machine learning methods validated the utility of \TADASHI{} and the approach in general, they also revealed the vastness of the search space: some kernels in real applications translated into hundreds of polyhedral schedule tree nodes, each of which could be modified by a combination of numerous transformations and parameters. This realization, in turn, necessitates future work to create a solution that would allow \TADASHI{} to run at the scale of large HPC clusters, enabling the discovery of polyhedral optimizations that go beyond those described in the literature. We leave this exploration for future work.


\section{Acknowledgement} ChatGPT~\cite{openaichatgpt} was used for Grammar and Spelling correction.

\begin{acks}
  We are eternally grateful to Sven (skimo) Verdoolaege for his invaluable help by replying to our (often silly) questions on the ISL mailing list.
\end{acks}

\bibliographystyle{ACM-Reference-Format}
\bibliography{emil-zotero,references,refs_ML}

\end{document}